\documentclass[review]{elsarticle}
\usepackage{lineno,hyperref}
\usepackage{wasysym}
\usepackage{graphicx}
\usepackage{adjustbox,lipsum}
\usepackage{subfig}
\usepackage{epstopdf}
\usepackage{multicol}
\usepackage{multirow}
\usepackage[inline]{enumitem}
\usepackage{siunitx}
\usepackage{threeparttable}
\usepackage{longtable}
\usepackage{epstopdf}
\usepackage{adjustbox}
\modulolinenumbers[5]
\usepackage{soul}
\usepackage[textsize=tiny,
textwidth=3.5cm]{todonotes}
\makeatletter
\if@todonotes@disabled

\else

\fi
\makeatother

\journal{Journal of \LaTeX\ Templates}

\begin{document}

\begin{frontmatter}

\title{Emotion Recognition in Low-Resource Settings: An Evaluation of Automatic Feature Selection Methods}
\author{Fasih Haider}
\address{Usher Institute, Edinburgh Medical School, the University of Edinburgh, UK}
\author{Senja Pollak}
\address{Jozef Stefan Institute, Ljubljana, Slovenia and Usher Institute, Edinburgh Medical School, the University of Edinburgh, UK}

\author{Pierre Albert}
\address{Usher Institute, Edinburgh Medical School, the University of Edinburgh, UK}

\author{Saturnino Luz}
\address{ Usher Institute, Edinburgh Medical School, the University of Edinburgh, UK}
\begin{abstract}
  Research in automatic affect recognition has seldom addressed the
  issue of computational resource utilization. With the advent of
  ambient intelligence technology which employs a variety of
  low-power, resource-constrained devices, this issue is increasingly
  gaining interest. This is especially the case in the context of
  health and elderly care technologies, where interventions may rely
  on monitoring of emotional status to provide support or alert carers
  as appropriate. This paper focuses on emotion recognition from
  speech data, in settings where it is desirable to minimize memory
  and computational requirements. Reducing the number of features for
  inductive inference is a route towards this goal. In this study, we
  evaluate three different state-of-the-art feature selection methods:
  Infinite Latent Feature Selection (ILFS), ReliefF and Fisher
  (generalized Fisher score), and compare them to our recently
  proposed feature selection method named `Active Feature Selection'
  (AFS).  The evaluation is performed on three emotion recognition
  data sets (EmoDB, SAVEE and EMOVO) using two standard acoustic
  paralinguistic feature sets (i.e. eGeMAPs and emobase). The results
  show that similar or better accuracy can be achieved using subsets
  of features substantially smaller than the entire feature set. A
  machine learning model trained on a smaller feature set will reduce
  the memory and computational resources of an emotion recognition
  system which can result in lowering the barriers for use of health
  monitoring technology.
\end{abstract}

\begin{keyword}
Feature Engineering, Feature Selection, Emotion Recognition, Affective Computing, Prosodic analysis, Cognitive Health Monitoring


\end{keyword}

\end{frontmatter}


\section{Introduction}
\label{sec:intro}
Speech signals are used in a number of automatic prediction tasks,
including cognitive state detection \cite{haider2015}, cognitive load
estimation \cite{bib:SchullerSteidlEtAl14in}, presentation quality
assessment \cite{haider2016ICASSP} and emotion recognition
\cite{el2011survey,bib:SchullerBatlinerEtAl11r}.
Emotional/affective states could have influence on health  and intervention outcomes. Positive emotions have been linked with health improvement, while negative emotions may have negative impact \cite{consedine_role_2007}.
For example, long term bouts of negative emotions are predisposing
factors for depression (ibid.), while positive emotions-related humour
and optimism have been linked with positive effects on the immune system and cardiovascular
health \cite{dimsdale_psychological_2008}.
Emotion recognition has been used in applications in the domain of
health technologies, including mental health assessment
and beyond \cite{valstar_avec_2013, desmet_emotion_2013,haider2020Lrec}.

Applications using speech usually extract emotions as an additional
signal in complex systems, such as in ambient intelligence (AmI)
\cite{mano_exploiting_2016}, depression recognition
\cite{desmet_emotion_2013}, and longitudinal cognitive status
assessment \cite{su2016predicting}.  These approaches employ very
high-dimensional feature spaces consisting of large numbers of
potentially relevant acoustic features, usually obtained by applying
statistical functionals to basic, energy, spectral and voicing related
acoustic descriptors \cite{eyben2009openear} extracted from speech
intervals lasting a few seconds
\cite{bib:VerveridisKotropoulos06em}. Although there is no general
consensus on what the ideal set of features should be, this
``brute-force'' approach of employing as many features as possible
seems to outperform alternative (Markovian) approaches to modelling
temporal dynamics on the classifier level
\cite{bib:WeningerEybenEtAl13acemaud}. However, the use of such
high-dimensional data sets poses challenges for prediction, as they
suffer from the so-called ``curse of dimensionality'', high degree of
redundancy in the feature set, and a large number of features with
poor descriptive value. Su and Luz, for instance, noted that in a
cognitive load prediction data set about 4\% of a feature set of over
250 features had a standard deviation of less than 0.01 and therefore
contributed negligibly to the classification task
\cite{su2016predicting}. Moreover, processing of very large numbers of 
features presents computational challenges for the
low-power, low-cost devices such as
the Raspberry Pi
Zero~\footnote{https://www.raspberrypi.org/products/raspberry-pi-zero/
  (last accessed January 2019)}, which are often used in AmI applications.

The main contribution of this study is the evaluation of different
state of the art feature selection methods, including our Active
Feature Selection (AFS) method, on the emotion recognition from
speech, which has, to the best of our knowledge, not yet been
systematically explored. This study extends our previous work
\cite{haider2018saameat}, where we first introduced the novel AFS
method and tested it on the ICMI Challenge on Eating Conditions Recognition
\cite{schuller_interspeech_2015}.

\section{Background and Related Work}
The automatic identification of emotions in speech is a challenging task, and identifying relevant acoustic features and systematic comparative evaluations have been difficult \cite{anagnostopoulos_features_2015}. In 2016, the eGeMAPs set \cite{eyben2016geneva} (see Section~\ref{sec:Feature Extraction}) was designed  
based on features' potential to reflect affective processes (extensively used in the literature) 
and their theoretical significance. It was proposed to set a common ground of emotion-related speech features, and it has become since then a \textit{de-facto} standard.
The set of target emotions has mostly been fixed around the `Big Six'
, and similarly, evaluations are more and more frequently performed on a number of publicly available corpora (see Section~\ref{subsec:data set}).
In the health domain, feature selection methods for speech processing have been applied to determine the most discriminant features in support of automatization efforts, e.g. for the assessment of patients with pre-dementia and Alzheimer's disease  \cite{konig_automatic_2015-1,8910399} or for the detection of sleep apnea  \cite{goldshtein_automatic_2011-1}.
The automatic emotion recognition problem gained a lot of attention in the past few years \cite{dhall2018emotiw,EmotiW2017} and is addressed by processing the facial, speech, body movements and biometric information of humans \cite{knyazev2017convolutional, Haider:2016:ARV:3011263.3011270,madzlan2015automatic,ICMI2017:HuEtAl, haider2015}.
Numerous studies \cite{knyazev2017convolutional, Haider:2016:ARV:3011263.3011270, ICMI2017:HuEtAl, Vielzeuf_Pateux_Jurie_2017,  Wang_ICMI2017, ouyang2017audio} extract audio features with OpenSMILE using \textit{de-facto} standard presets: IS10, GeMAPS, eGeMAPs, Emobase.

The reviewed literature suggests that although the accuracy of various machine learning approaches in this area is promising, automatic dimensionality reduction has focused largely on the removal of noisy or redundant features, with less attention paid to computational resource utilisation \cite{ haider2015,knyazev2017convolutional, Haider:2016:ARV:3011263.3011270,madzlan2015automatic,ICMI2017:HuEtAl, Vielzeuf_Pateux_Jurie_2017,  Wang_ICMI2017, ouyang2017audio}.

There are many dimensionality reduction methods: some are feature selection methods which require labelled data such as correlation based feature selection and Fisher feature selection \cite{hall1999correlation,gu_generalized_2012}, and some are feature transformation methods which do not require labelled data such as Principle Component Analysis (PCA), independent component analysis  \cite{wang2006independent}. Recently, efforts have been spent to reduce dimensionality using PCA to improve the results for emotion recognition from speech \cite{jagini2017exploring, aher2016analysis, wang2010speech,Haider2018} in different settings such as noisy setting \cite{aher2016analysis} but dimensionality reduction using feature selection methods is less explored. 

\section{Feature Selection Methods}
\label{Feature Selection Methods}
In this section we will briefly describe the feature selection methods used in this study along with our AFS method. We have selected three state of the art feature selection methods and motivation behind using them is their robust performance as demonstrated by Roffo et. al \citep{roffo2017infinite}.

\subsection{Infinite Latent Feature Selection (ILFS)}
The ILFS method \citep{roffo2017infinite} performs cross-validation on unsupervised ranking of each feature.
In a pre-processing stage, each feature is represented by a descriptor reflecting how discriminative it is.
A probabilistic latent graph containing each feature is built. Weighted edges model pairwise relations among feature distributions, created using probabilistic latent semantic analysis. The relevance  
of each feature is computed by looking on its weight in arbitrary set of cues.
Each path in the graph represents a selection of features. The final ranking of each feature looks at its redundancy in all the possible feature subsets, selecting the most discriminative and relevant features. 
The evaluation on a range of different tasks, e.g. object recognition classification and DNA Microarray, confirms its robustness, outperforming other methods on robustness and ranking quality \citep{roffo2017infinite}.

\subsection{ReliefF}
The ReliefF algorithm \cite{robnik-sikonja_adaptation_1997} which is an adaptation of Relief \cite{kira1992feature}, performs a ranking and selection of top scoring features based on their processed score. 
The score is calculated by weighting features on a random sample of instances. For each instance, the weight vector represents the relevance of each feature amongst the class labels: neighbours are selected from the same class (nearest hits) and from each different class (nearest misses). The weight of each feature increases when the difference with its nearest hits is low and with its nearest misses is high. Each weight vector is combined in a global relevance vector.
The final subset is constituted of all the features with relevance greater than a manually set threshold.
ReliefF is a common method of Feature Selection which has been continuously improved since its first publication  \cite{kira1992feature,robnik-sikonja_adaptation_1997}.
 

\subsection{Generalized Fisher score (Fisher)}
The generalized Fisher score \cite{gu_generalized_2012} is a generalization of the Fisher score to take into account redundancy and combination of features.
A subset of features is found to maximize the lower bound of traditional Fisher score. The combination of features is evaluated, and redundant features discarded. A quadratically
constrained linear programming (QCLP) is solved with a cutting plane algorithm. In 
each iteration, a multiple kernel learning is solved by a multivariate ridge regression followed by a projected gradient descent to update the kernel weights.
The method produces state of the art results, outperforming many feature selection methods while having a lower complexity \cite{gu_generalized_2012}.

\subsection{Active feature selection method}
\label{sec:AFS selection}
\begin{figure}[htb]
	\centering
	\centerline{\includegraphics[width=1\linewidth]{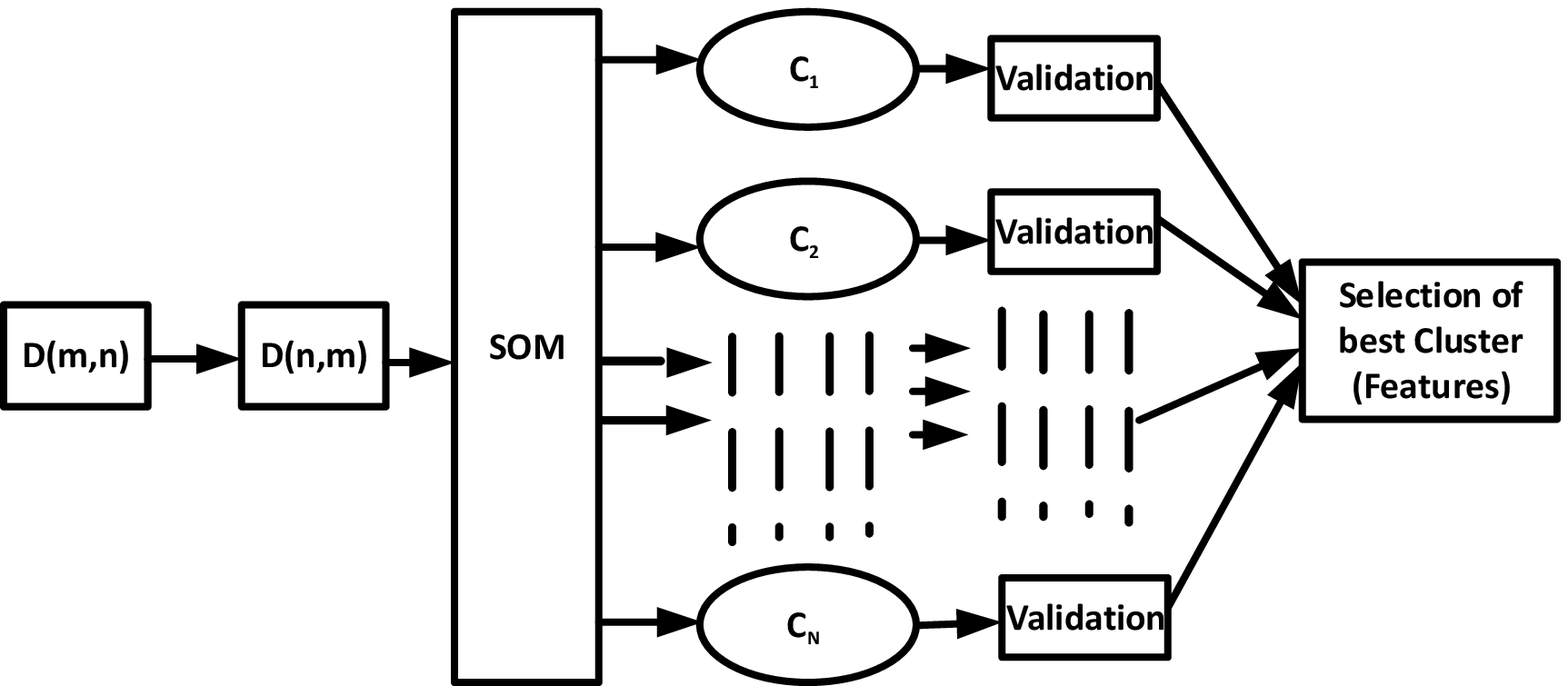}}
	\caption{ Active feature selection method: D(m,n) represents the data where \emph{m} is the total number of training instances and \emph{n} is the total number of dimensions (988 for \textit{emobase} and 88 for \textit{eGeMAPs}) \cite{haider2018saameat}. }
    \label{fig:AFSSystem}  
\end{figure}

An Active Feature Selection method, which divides a feature set into subsets, has been recently introduced \cite{haider2018saameat}. The term `Active' is used because compared to other approaches it evaluates feature subsets and not each feature separately  meaning that different features actively contribute to the feature selection. We are not clustering the number of instances but the dimensions. Our hypothesis is that the noisy features have  different characteristics than informative features, and that clustering the features will divide the features into many subsets according to their common characteristics. It involves clustering the data set into N clusters (where $N=5,10,15, ..., 100$) using Self-Organizing Maps (SOM) with 200 iterations and batch training \cite{kohonen1998self}, and then evaluating discrimination power of features present in each cluster $C_{N}$  using Leave-One-Subject Out (LOSO) cross-validation setting, as depicted in Figure~\ref{fig:AFSSystem}, and selecting the cluster with the highest validation accuracy  (see Figure~\ref{fig:hits} in Section~\ref{Results and discussion}).

\if comment
\hl{Suggestion for rewriting the previous paragraph is in comment, please decide which one you prefer:}
\label{sec:AFS selection}
An Active Feature Selection method, which divides a feature set into subsets, has been recently introduced \cite{haider2018saameat}. The term `Active' is used because compared to other approaches it evaluates feature subsets and not each feature separately.\hl{SP: for me, this explanation is not fully transparent, but it is not a problem. If there is an easy extra link, please add it, e.g. ... and not each feature separately, meaning that different features actively contribute to the feature selection.} We are not clustering the number of instances but the dimensions. Our hypothesis is that the noisy features have  different characteristics than informative features, and that clustering the features will divide the features into many subsets according to their common characteristics. It involves clustering the data set into N clusters (where $N=5,10,15, ..., 100$) using Self-Organizing Maps (SOM) with 200 iterations and batch training \cite{kohonen1998self}, and then evaluating discrimination power of features present in each cluster $C_{N}$  using Leave-One-Subject Out (LOSO) cross-validation setting, as depicted in Figure \ref{fig:AFSSystem}, and selecting the cluster with the highest validation accuracy  (see Figure~\ref{fig:hits} in Section XXXADD CROSS-REF).
\fi

\section{Experimentation}
The section describes the datasets and their characteristics along with acoustic feature extraction and classification methods. 
\subsection{Data sets}
\label{subsec:data set}
Three corpora were selected for their shared characteristics and public availability: EmoDB, SAVEE, and EMOVO. They consist of recorded acted performances, annotated using the well-known and widely used \textit{Big Six} set of annotations : anger, disgust, fear, happiness, sadness, surprise + neutral, except in the older EmoDB data set where boredom was used instead of surprise. Their characteristics are summarised in Tables \ref{table:chardb} and \ref{table:repdb}.

\textit{Berlin Database of Emotional Speech (EmoDB)}\\
The EmoDB corpus \cite{burkhardt_database_2005} is a data set commonly used in the automatic emotion recognition literature. It features 535 acted emotions in German, based on utterances carrying no emotional bias. The corpus was recorded in a controlled environment resulting in high quality recordings. 
Actors were allowed to move freely around the microphones, affecting absolute signal intensity.
In addition to the emotion, each recording was labelled with phonetic transcription using the SAMPA phonetic alphabet, emotional characteristics of voice, segmentation of the syllables, and stress.
The quality of the data set was evaluated by perception tests carried out by 20 human participants. In a first recognition test, subjects listened to a recording once before assigning one of the available categories, achieving an average recognition rate of 86\%. A second naturalness test was performed. Documents achieving a recognition rate lower than 80\% or a naturalness rate lower than 60\% were discarded from the main corpus, reducing the corpus to 535 recordings from the original 800.

\textit{Surrey Audio-Visual Expressed Emotion (SAVEE)}\\
SAVEE \cite{HaqJackson_AVSP09} is an audio-visual data set that was recorded to support the development of an automatic emotion recognition system. The corpus is a set of 480 British English utterances. 
Each actor was recorded for 15 utterances per emotion
 (3 common utterances \- recorded for each of the 7 emotions, 2 emotion specific, and 10 generic sentences \- different for each emotion) and 30 neutral recordings (the 3 common and every emotion specific sentences). 
No limitation regarding audio features (e.g. absolute signal intensity) is explicitly stated in the description of the data set.
A qualitative evaluation of the database was run as a perception tests by 10 human subjects. The mean classification accuracy for the audio modality was 66.5\%, 88\% for the visual modality, and 91.8\% for the combined audio-visual modalities.

\textit{Italian Emotional Speech Database (EMOVO)}\\
The EMOVO corpus \cite{costantini_emovo_2014} is a speech data set featuring recorded emotions from acted performances by 6 persons. 
Actors were allowed to move freely around the microphones and the volume was manually adjusted, affecting absolute signal intensity.
A qualitative evaluation was performed using a discrimination test. Two phrases were selected and, for each, 12 subjects had to choose between two proposed emotions. The mean accuracy for the test was about 80\%.

\begin{table}[!htb]
\caption{Main characteristics of the data sets.}
\begin{tabular}{l  m{1.4cm}  m{2cm}  m{2cm} m{3cm}}
  \hline
	Corpus & Size (utterances) & Population & Participants & Emotion categories\\
	\hline
	EmoDB & 535 & 10 (5 males, 5 females) & German native speakers actors & anger, disgust, fear, joy, sadness, \textit{boredom} + neutral\\[2ex]
	SAVEE & 480 & 4 (males) & English native speakers actors & anger, disgust, fear, happiness, sadness, \textit{surprise} + neutral\\[2ex]
	EMOVO & 588 & 6 (3 males, 3 females) & Italian native speakers actors & anger, disgust, fear, happiness, sadness, \textit{surprise} + neutral\\
  \hline
\end{tabular}

\label{table:chardb}
\end{table}

\begin{table}[!htb]
\caption{Distribution of recordings across emotion categories.}
\resizebox{\columnwidth}{!}{%
\begin{tabular}{l  c  c  c  c  c  c  c  c}
\hline
Corpus & Neutral & Anger & Disgust & Fear & Happiness & Sadness & Surprise & Boredom\\
\hline
EmoDB & 79 & 127 & 46 & 69 & 71 & 62 & - & 81\\
SAVEE & 120 & 60 & 60 & 60 & 60 & 60 & 60 & -\\
EMOVO & 84 & 84 & 84 & 84 & 84 & 84 & 84 & -\\
\end{tabular}
}
\label{table:repdb}
\end{table}

\subsection{Volume normalization and feature extraction}
\label{sec:Feature Extraction}
We have normalized all the speech utterances' volume in to the range [-1:+1] dBFS before any acoustic feature extraction.
The motivation of doing this normalization is to make the model robust against different recording conditions such as distance between microphone and subject. We use the openSMILE \cite{eyben2013recent} toolkit for the extraction of two acoustic feature sets which are widely used for emotion recognition as follow:  

\textbf{emobase:}
This acoustic feature set contains the MFCC,
voice quality, fundamental frequency (F0), F0 envelope, LSP
and intensity features along with their first and second order
derivatives. In addition, many statistical functions are applied
to these features, resulting in a total of 988 features for every
speech utterance.

\textbf{eGeMAPs:}The eGeMAPs
\cite{eyben2016geneva} feature set contains the F0
semitone, loudness, spectral flux, MFCC, jitter, shimmer, F1,
F2, F3, alpha ratio, hammarberg index and slope V0 features
including many statistical functions applied to these features, 
which resulted in a total of 88 features for every speech utterance.

\subsection{Classification Method}
\label{sec:Classification Methods}
The classification is performed using  
Support Vector Machines (SVM) using SMO solver with \textit{box constraint} (k) of 0.75 and linear kernel function.  This classifier is employed in
MATLAB\footnote{http://uk.mathworks.com/products/matlab/ (Last accessed: January 2019)} using the statistics and machine learning
toolbox. The feature selection methods are evaluated in LOSO cross-validation setting for SVM classifier using Unweighted Average Recall (UAR) measure.

\subsection{Evaluation Criteria}
All of the emotion recognition data sets are labeled for seven classes and we have evaluated the classifier using UAR which is the average accuracy of all classes. The UAR measure is selected because the datasets are not balanced for emotions. The method with the highest UAR is considered the best. The blind/majority guess for this task is the UAR of 14.3\%.
However, as our focus is on feature selection methods, we set the baseline as UAR obtained using the entire feature set.

\section{Results and discussion}
\label{Results and discussion}
\begin{figure}[!htb]
	\centering
	\centerline{\includegraphics[width=0.95\linewidth]{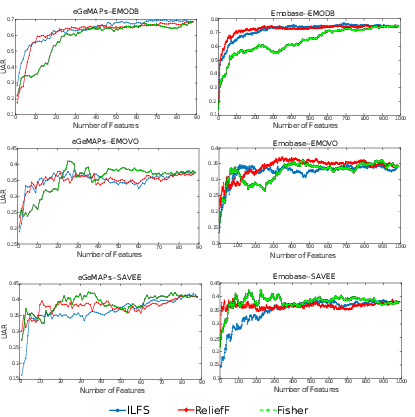}}
	\caption{Feature selection methods (ILFS, ReliefF and Fisher) results for all three data sets (EMODB, EMOVO and SAVEE) using two feature sets (eGeMAPs and emobase). Where x-axis represents the number of features and y-axis represents the UAR.}
    \label{fig:FS}  
\end{figure}

We have evaluated the three different automatic feature selection methods named ILFS, ReliefF and Fisher along with our newly proposed AFS method using two different acoustic feature sets extracted from three different data sets. The results of three feature selection methods are shown in Figure~\ref{fig:FS}~\footnote{The AFS results are not plotted as the method is not directly comparable given that AFS is not about ranking a single feature but a subset of features.}. It can be observed that around 30 out of 88 eGeMAPs features and around 100 out of 988 emobase features are sufficient in providing almost the same UAR as the highest achieved UAR for three data sets. 
The best results of each feature selection method are depicted in Table~\ref{tab:best}. 
\begin{table}[!htb]
\centering
\caption{Best Unweighted Average Recall (UAR (\%)) of feature selection methods and number of selected features (numFeat) are reported. The best UAR (\%) results for each feature set are given in bold. The unweighted arithmetic average for each feature selection method is also reported in `Average' column.}
\begin{adjustbox}{width=1\textwidth}
\begin{tabular}{c|cc|cc|cc|cc|cc|cc|c}
 Data Set & \multicolumn{4}{c}{EmoDB} & \multicolumn{4}{c}{EMOVO} & \multicolumn{4}{c}{SAVEE} & \multicolumn{1}{|c}{Average} \\ \hline
Feature Set& \multicolumn{2}{c}{ eGeMAPs} & \multicolumn{2}{c|}{emobase} & \multicolumn{2}{c}{ eGeMAPs} & \multicolumn{2}{c|}{emobase} & \multicolumn{2}{c}{ eGeMAPs} & \multicolumn{2}{c}{emobase} &  \multicolumn{1}{|c}{--}\\ \cline{1-13}
         & numFeat   & UAR (\%)      &   numFeat &   UAR (\%)     & numFeat   &    UAR   & numFeat   & UAR(\%)       &  numFeat  & UAR (\%)       & numFeat   & UAR (\%)       & -- \\
Baseline &  88       & 68.5    &  988      &  74.6   &    88     &    37.4 & 988       & 34.4    &  88       & 40.8    &  988      &  38.1   &  49.0 \\
    ILFS &  74     & \textbf{69.7}    &  685      &  \textbf{76.9 }  &    28     &    38.1 & 113       & 34.7    &  86       & 42.0    &   574     &  38.8   &  46.9 \\
 reliefF &  88       & 68.5    &  666      &  75.3   &    20     &    37.8 & 348       & \textbf{37.1}    &  82       & 41.4    &  72       &  39.3   &  49.9 \\
  Fisher &  88       & 68.5    &  975      &  75.2   &    25     &    \textbf{41.0}  & 464       & 36.2    &  34       & \textbf{42.4}    &  158      &   \textbf{42.4}  &  \textbf{51.0} \\
     AFS &  81       & 68.5    &  696      &  75.8   &     2     &    39.0 & 56        & 36.4    &  68       & 40.5    &   21      &   37.5  &  49.6 
\end{tabular}\label{tab:best} 
\end{adjustbox}
\end{table}

The results confirm that a higher accuracy can be achieved using a subset of the feature set than when using the full feature set. The results for each data set are as follows:

\begin{enumerate}

   \item  EmoDB:  ILFS method provides better UAR (69.7\% for eGeMAPs and 76.9\% for emobase) results than other methods and is  
   able to reduce the number of features (74 out of 88 for eGeMAPs and 685 out of 988 for emobase). The confusion matrix of the best UAR (76.9\%) is shown in Figure~\ref{fig:ilfsEmobaseEmoDB}. For eGeMAPs, AFS method provides an UAR of 68.5\% (around 1\% lower than ILFS) using 81 features. For emobase, AFS method provides an UAR of 75.8\% (around 1\% lower than ILFS) using 696 features. With a subset of eGeMAPs feature set, the reliefF and Fisher methods are not able to improve over the baseline in terms of UAR. 
   However,  Figure~\ref{fig:FS} shows that reliefF and Fisher achieved almost the same UAR as compared to baseline  with only 35 eGeMAPs features instead of 88 eGeMAPs features. Hence around 60\% reduction in number of features is observed.    
\begin{figure}[!htb]
	\centering
	\centerline{\includegraphics[width=0.65\linewidth]{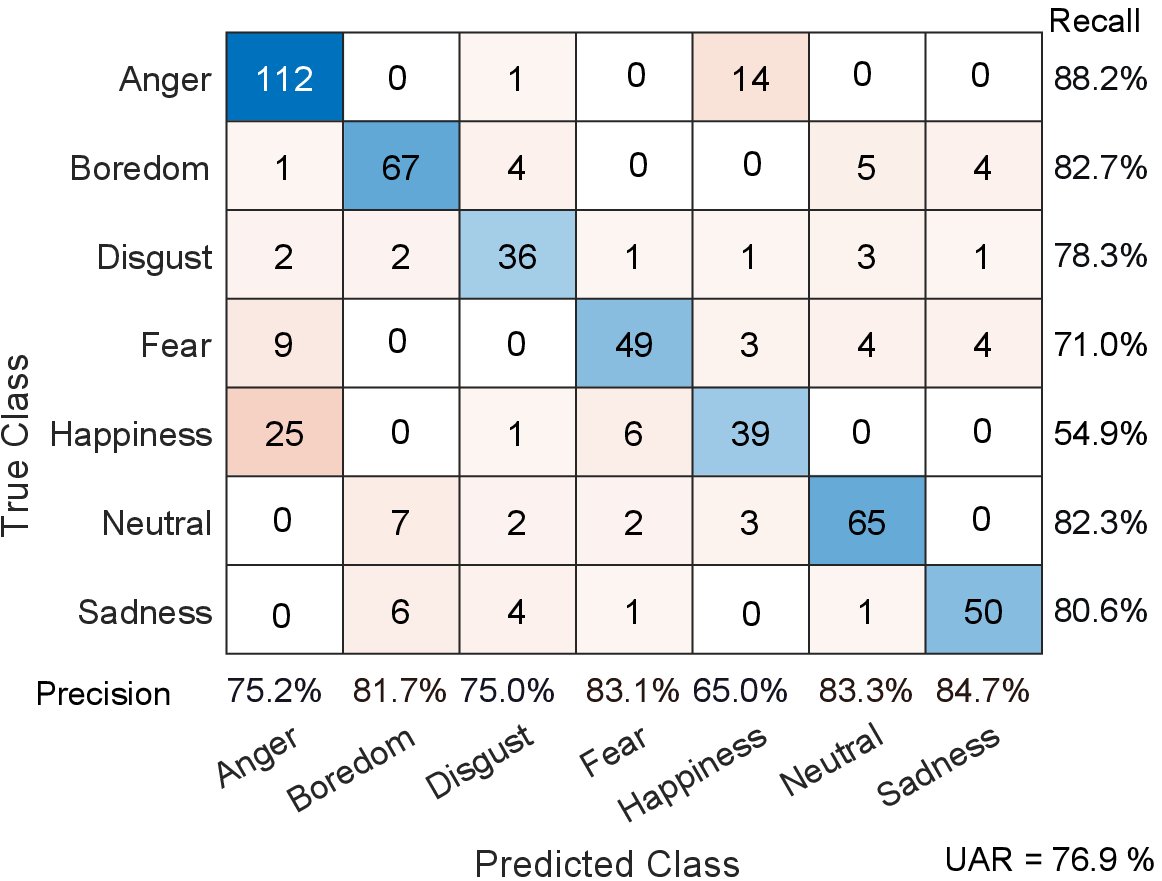}}
	\caption{Confusion matrix of ILFS Feature selection method  for EmoDB data set using emobase feature set.}
    \label{fig:ilfsEmobaseEmoDB}  
\end{figure}

   \item  EMOVO: Fisher method provides better UAR (41.0\%) than other methods for eGeMAPs feature set (using only 25 out of 88 features) and ReliefF method provides a better UAR (37.1\%) than other methods for emobase feature set (348 out of 988).  The confusion matrix of the best UAR (41.0\%) is shown in Figure~\ref{fig:fisherGemapsEMOVO}. The results of AFS are slightly lower than the best method (for around 2 \%), but the  number of features are significantly lower, compared to other methods. Note that AFS selects only 2 eGeMAPs features out of 88 and 56 emobase feature out of 988  which provides the UAR of 39.0\% 
   and 36.4\%, respectively.   
   
   \begin{figure}[!htb]
	\centering
	\centerline{\includegraphics[width=0.65\linewidth]{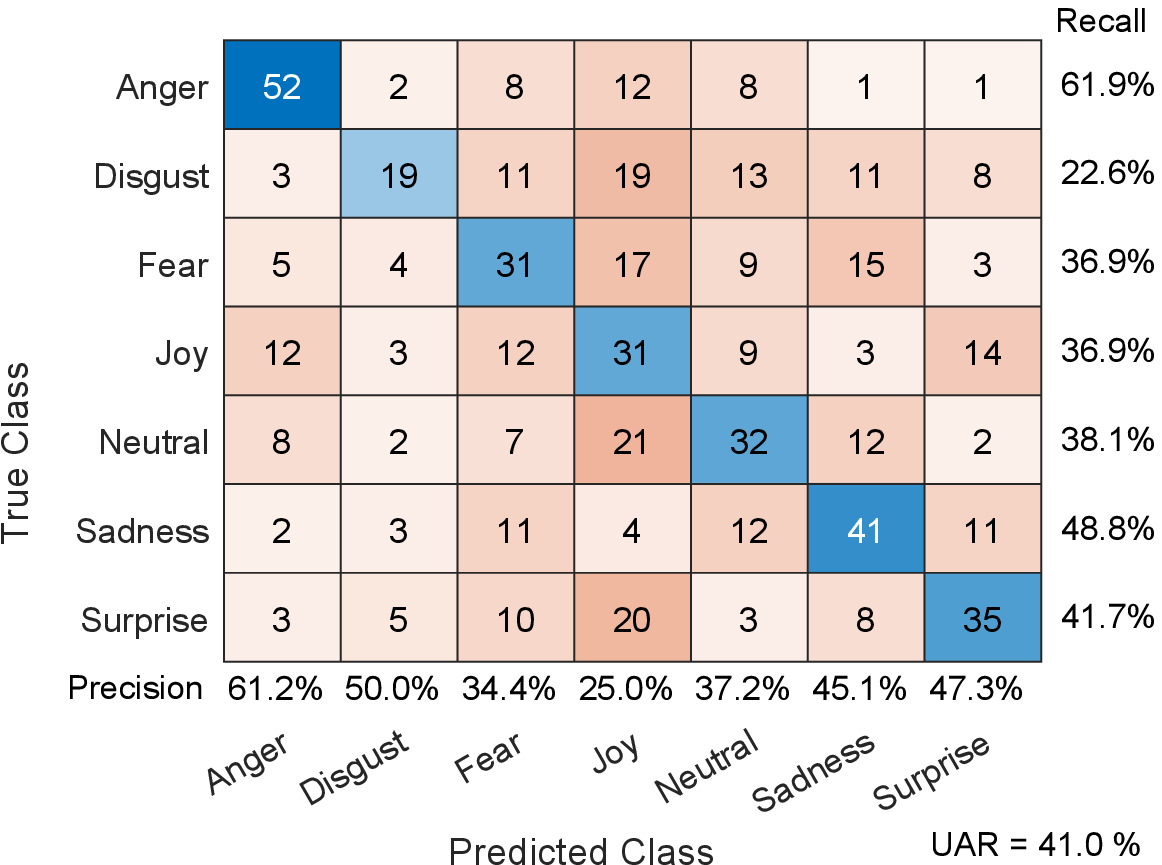}}
	\caption{Confusion matrix of Fisher feature selection method for EMOVO data set using eGeMAPs feature set.}
    \label{fig:fisherGemapsEMOVO}  
\end{figure}

   \item SAVEE:  Fisher method provides better UAR than other methods for eGeMAPs feature set (34 out of 88 with a UAR of 42.4\%)  and emobase feature set (158 out of 988 with a UAR of 42.4\%). The confusion matrix of the best result (UAR = 42.4\%) using eGeMAPs features is shown in Figure~\ref{fig:fisherGemapsSAVEE}. For eGeMAPs, the results of AFS are slightly lower than the best method (around 2\%).  For emobase, AFS method provides an UAR of 37.5\% (around 5\% lower than Fisher) using 21 features.

   \begin{figure}[!htb]
	\centering
	\centerline{\includegraphics[width=0.65\linewidth]{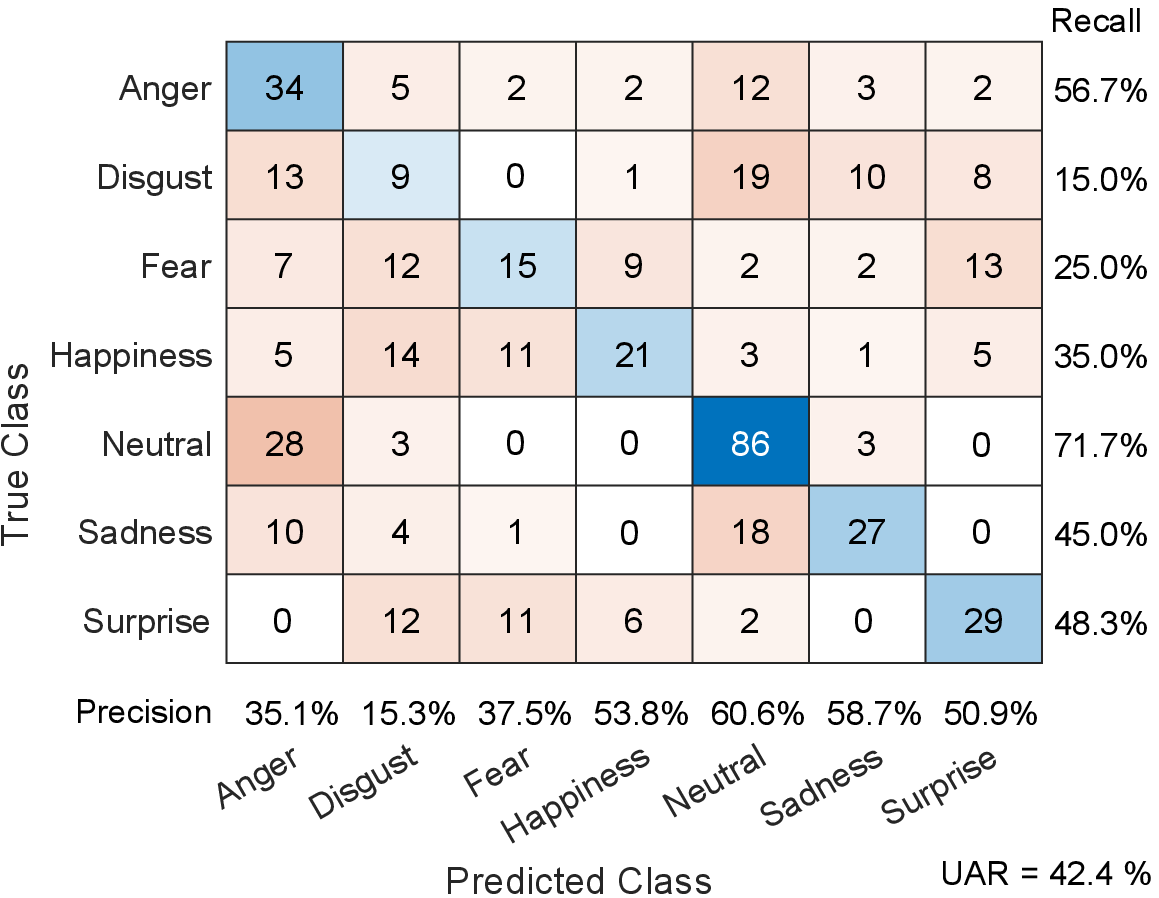}}
	\caption{Confusion matrix of Fisher feature selection method for SAVEE data set using eGeMAPs feature set.}
    \label{fig:fisherGemapsSAVEE}  
\end{figure}
   
\end{enumerate}

The machine learning models trained using EmoDB (76.9\%) data  provide 
better UAR than EMOVO (41.0\%) and SAVEE (42.9\%). This could be due to very high quality nature of the EmoDB data set. The EmoDB data set quality is evaluated by 20 human subjects with an average recognition rate of 86\%, and
the audio recording with the inter-coder agreement below 80\% were removed (no such measure was taken for EMOVO and SAVEE). 
For EMOVO, even if the reported accuracy for the test set is 80\% (see Section~\ref{subsec:data set}), one should note that there, instead of evaluating full EMOVO data set only two phrases were selected and a subject had to choose only between two proposed emotions (instead of seven emotions). However, our machine learning approach on EMOVO data is seven class classification problem, which explains lower results. 
For SAVEE, 10 human subjects evaluated the data set and came up with an accuracy of 66.5\% for audio. Our machine learning based models provide promising results as compared to humans subjects. Although they are less accurate than human annotators, we use only acoustic information to automate the process of emotion recognition, while human annotators used both acoustic and linguistic information (i.e. the spoken content). 

\begin{figure}[!htb]
	\centering
	\centerline{\includegraphics[width=0.75\linewidth]{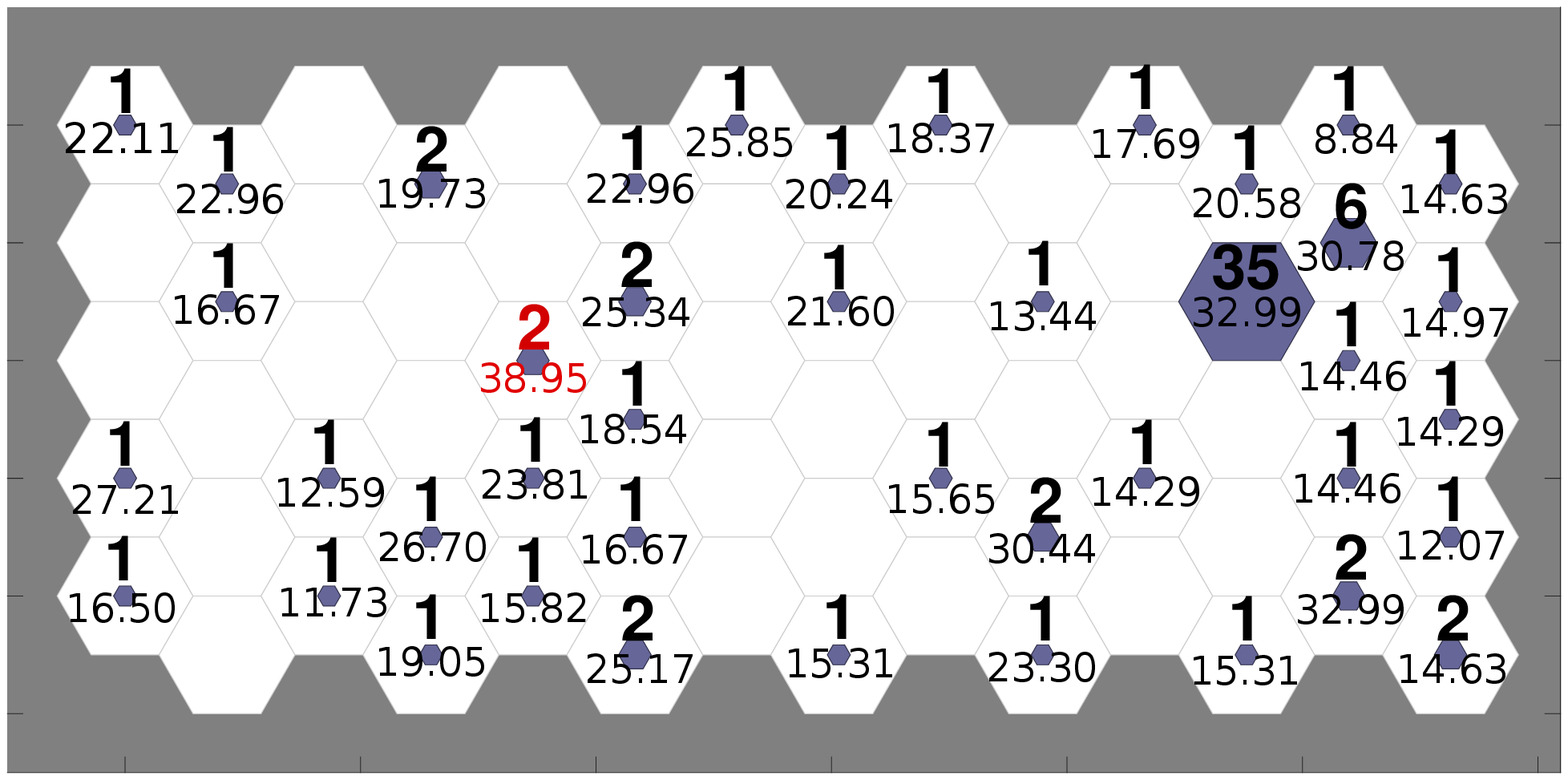}}
	\caption{A visualization of AFS method results: number of features present in each cluster (i.e. hexagon or neuron) along with the UAR (\%) obtained using eGeMAPs feature set for EMOVO data set. Note that 
	2 out of 88 features provide 
	better results than other feature subsets.}
    \label{fig:hits}  
\end{figure}

\begin{figure}[!htb]
	\centering
	\centerline{\includegraphics[width=0.95\linewidth]{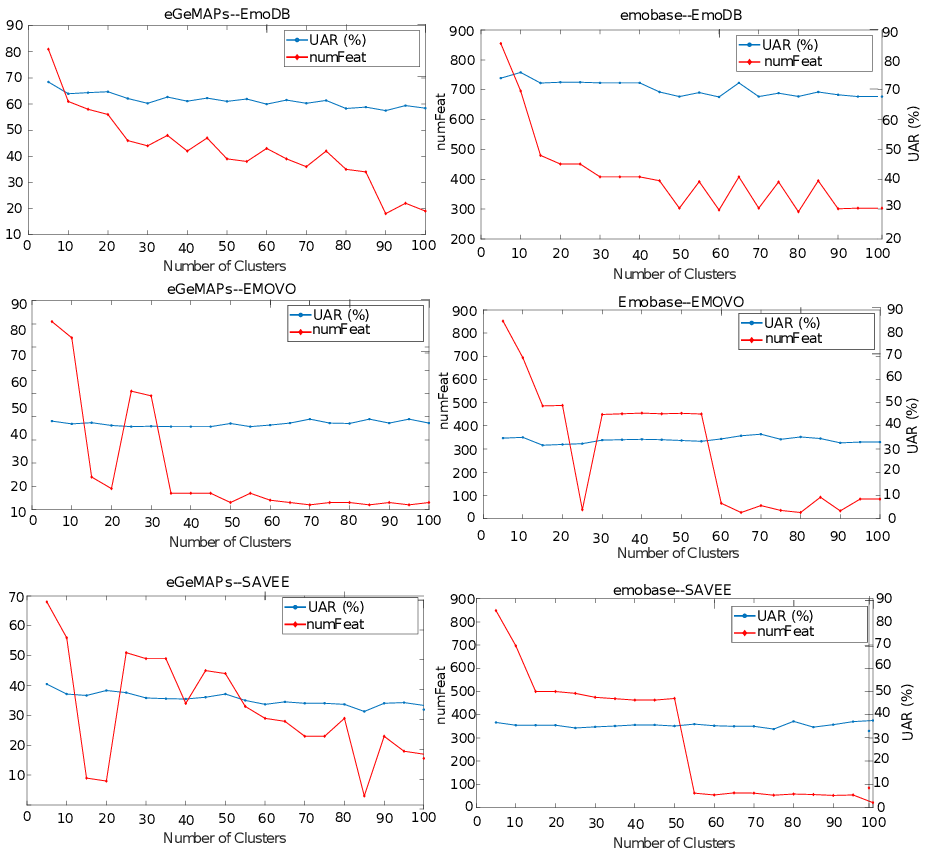}}
	\caption{AFS method results: The x-axis represents the number of cluster (N = 5,10,15, ... 100). The y-axis represents the number of features (numFeat) and Unweighted Average Recall (UAR) in \% of the best cluster.}
    \label{fig:AFS}  
\end{figure}

As shown in Table~\ref{tab:best}, Generalized Fisher score provides better results in 3 out of 6 cases,
ILFS provides better results in 2 out of 6 cases and reliefF provides better results in 1 out of 6 cases, indicating that overall Fisher feature selection provides the best results for the emotion recognition task.

The AFS method comes second in 3 out of 6 cases as shown in Table~\ref{tab:best}. It is also observed that the AFS method provides almost the same results in terms of UAR as the other state of the art feature selection method, with a smaller number of dimensions. 
We have noticed that for the SAVEE data set only 2 out of 88 eGeMAPs features (selected by AFS) provide better results than reliefF, ILFS and the baseline (i.e. entire feature set) and to get further insight of results, we demonstrate the evaluation of clusters (feature subsets) using AFS as shown in Figure~\ref{fig:hits}.  From Figure~\ref{fig:hits} we observe that there are many clusters (of features) which provide better results than the blind guess (14.3\%), while the feature cluster selected by AFS is containing only 2 features (i.e. \textit{hammarbergIndexV\_sma3nz\_amean} and \textit{hammarbergIndexV\_sma3nz\_amean}) and leads to the 39.0\% UAR. One of the possible lines of future work is to combine 
the features from different clusters to see if it leads to any improvement in classification in terms of UAR. AFS method is also evaluated with different number of clusters for SOM algorithm. The best UAR is obtained using 70 cluster for EMOVO dataset and UAR for all 70 clusters 
wit number of features (numFeat)  are shown in Figure~\ref{fig:hits}.

To further evaluate the feature selection methods, we have combined all three data sets which results in a 8-class problem i.e. to recognise (7+1) emotions. 
The results of this experimentation in LOSO cross-validation setting is shown 
in Table~\ref{tab:all}. We have noted that the reliefF method provides the best results for eGeMAPs (46.6\%) and emobase (48.0\%) feature sets. All three data sets belong to different languages and have different qualities of annotation. Hence, the reliefF method could be a better choice than other methods where the quality and language of data sets are different.

\begin{table}[!htb]
\centering 
\caption{Evaluation of feature selection methods for  7+1 emotion recognition task by combining all three data sets. Best Unweighted Average Recall (UAR (\%)) and number of selected features (numFeat) are reported. The bold figures indicate the best UAR (\%) for each feature set (i.e. eGeMAPs and emobase).} \footnotesize
\begin{tabular}{ccccc}\footnotesize
Method & \multicolumn{2}{c}{eGeMAPs} & \multicolumn{2}{c}{emobase} \\ \hline
 &     numFeat     &   UAR (\%)        &     numFeat      &  UAR (\%)        \\
Baseline &     88     &   44.4        &     988      &     47.4     \\
ILFS     &     78      &    45.6      & 709      &   47.9       \\
relifF   &   44        & \textbf{46.6}          &     732      & \textbf{48.0}        \\
Fisher   &   53        &  45.3        &      822     &     47.9     \\
AFS      &     79      &    43.8      &    835       &     47.2    
\end{tabular}
\label{tab:all}
\end{table}

In a previous study \cite{haider2018saameat}, we demonstrate that the AFS  method is able to select a feature subset  which provides better results than the entire feature set and the PCA feature set for eating condition recognition. However the results have not been demonstrated in detail as in this study and the AFS has not been evaluated on the multiple data sets and compared against other feature selection methods which is a step towards in demonstrating the generalisability of the AFS method.
The contribution of this study is not only the evaluation of performance of different feature selection methods
but also the assessment of the extent to which AFS, reliefF, Fisher and ILFS can reduce the feature set and therefore 
select small enough subsets which will impose lower computational demands on low resource systems, while preserving or improving emotion recognition performance, in comparison to full feature sets.

\section{Conclusion}
This study evaluates three `state of the  art' feature selection methods named infinite latent feature selection (ILFS), reliefF and generalized Fisher score for emotion recognition along with a recently proposed `Active Feature Selection' method. It utilizes three different emotion recognition data sets, namely EmoDB, EMOVO and SAVEE from three different languages i.e. German, English and Italian respectively. The results show that a higher UAR can be achieved using a subset of the full feature set. In summary, around 30 out of 88 eGeMAPs and 100 out of 988 emobase features are sufficient to obtain almost the same UAR as a full feature set.
The Fisher feature selection method provides the best averaged UAR of 51.0\% across all three data sets and two feature-sets compared to the 49.0\% averaged UAR for baseline method (without feature selection). However the reliefF method outperforms all the methods when all the data sets have been combined for emotion recognition.
This finding is relevant to the development of machine learning models for machines with low computational resources. The AFS method provides competitive results 
compared to other methods. However, AFS currently uses only features present in one cluster. For future studies, we will explore methods to rank the clusters of features and do fusion of different clusters for possible accuracy improvements. Other possible avenues for future work include testing the AFS on other modalities along with speech. 

\section{Acknowledgement}
This research is funded by the European Union's Horizon 2020 research
program, under grant agreement No 769661, towards the SAAM
project. PA is supported by the Medical Research Council (MRC). The work of S. Pollak was partially supported by the Slovenian Research Agency (ARRS) core research programme \emph{Knowledge Technologies} (P2-0103). 

\section*{References}
\bibliography{CSLMain.bib}

\begin{thebibliography}{10}
\expandafter\ifx\csname url\endcsname\relax
  \def\url#1{\texttt{#1}}\fi
\expandafter\ifx\csname urlprefix\endcsname\relax\def\urlprefix{URL }\fi
\expandafter\ifx\csname href\endcsname\relax
  \def\href#1#2{#2} \def\path#1{#1}\fi

\bibitem{haider2015}
H.~Akira, F.~Haider, L.~Cerrato, N.~Campbell, S.~Luz, Detection of cognitive
  states and their correlation to speech recognition performance in
  speech-to-speech machine translation systems, in: Proceedings of the 16th
  Annual Conference of the International Speech Communication Association,
  INTERSPEECH 2015, International Speech Communications Association, 2015, pp.
  2539--2543.

\bibitem{bib:SchullerSteidlEtAl14in}
B.~Schuller, S.~Steidl, A.~Batliner, J.~Epps, F.~Eyben, F.~Ringeval, E.~Marchi,
  Y.~Zhang, The {INTERSPEECH} 2014 computational paralinguistics challenge:
  {Cognitive} \& physical load, in: Proceedings of the 15th Annual conference
  of the International Speech Communication Association, INTERSPEECH 2014,
  International Speech Communications Association, 2014, pp. 427--431.

\bibitem{haider2016ICASSP}
F.~Haider, L.~Cerrato, N.~Campbell, S.~Luz, Presentation quality assessment
  using acoustic information and hand movements, in: Proceedings of the IEEE
  International Conference on Acoustics, Speech and Signal Processing (ICASSP),
  Institute of Electrical and Electronics Engineers (IEEE), 2016, pp.
  2812--2816.

\bibitem{el2011survey}
M.~El~Ayadi, M.~S. Kamel, F.~Karray, Survey on speech emotion recognition:
  Features, classification schemes, and databases, Pattern Recognition 44~(3)
  (2011) 572--587.

\bibitem{bib:SchullerBatlinerEtAl11r}
B.~Schuller, A.~Batliner, S.~Steidl, D.~Seppi, Recognising realistic emotions
  and affect in speech: {State} of the art and lessons learnt from the first
  challenge, Speech Communication 53~(9–10) (2011) 1062--1087.

\bibitem{consedine_role_2007}
N.~S. Consedine, J.~T. Moskowitz, The role of discrete emotions in health
  outcomes: {A} critical review, Applied and Preventive Psychology 12~(2)
  (2007) 59--75.

\bibitem{dimsdale_psychological_2008}
J.~E. Dimsdale, Psychological stress and cardiovascular disease, Journal of the
  American College of Cardiology 51~(13) (2008) 1237--1246.

\bibitem{valstar_avec_2013}
M.~Valstar, B.~Schuller, K.~Smith, F.~Eyben, B.~Jiang, S.~Bilakhia,
  S.~Schnieder, R.~Cowie, M.~Pantic, {AVEC} 2013: the continuous audio/visual
  emotion and depression recognition challenge, in: Proceedings of the 3rd
  {ACM} international workshop on {Audio}/visual emotion challenge (AVEC),
  Association for Computing Machinery, 2013, pp. 3--10.

\bibitem{desmet_emotion_2013}
B.~Desmet, V.~Hoste, Emotion detection in suicide notes, Expert Systems with
  Applications 40~(16) (2013) 6351--6358.

\bibitem{haider2020Lrec}
F.~Haider, S.~{De La Fuente Garcia}, P.~Albert, S.~Luz, Affective speech for
  alzheimer's dementia recognition., in: D.~Kokkinakis, K.~{Lundholm Fors},
  C.~Themistocleous, M.~Antonsson, M.~Eckerstr{\"o}m (Eds.), LREC: Resources
  and ProcessIng of linguistic, para-linguistic and extra-linguistic Data from
  people with various forms of cognitive/psychiatric/developmental impairments
  (RaPID), European Language Resources Association (ELRA), 2020, pp. 67--73.

\bibitem{mano_exploiting_2016}
L.~Y. Mano, B.~S. Faiçal, L.~H. Nakamura, P.~H. Gomes, G.~L. Libralon, R.~I.
  Meneguete, P.~R. Geraldo~Filho, G.~T. Giancristofaro, G.~Pessin,
  B.~Krishnamachari, Exploiting {IoT} technologies for enhancing {Health}
  {Smart} {Homes} through patient identification and emotion recognition,
  Computer Communications 89 (2016) 178--190.

\bibitem{su2016predicting}
J.~Su, S.~Luz, Predicting cognitive load levels from speech data, in: Recent
  Advances in Nonlinear Speech Processing, Springer, 2016, pp. 255--263.

\bibitem{eyben2009openear}
F.~Eyben, M.~W{\"o}llmer, B.~Schuller, Openear—introducing the munich
  open-source emotion and affect recognition toolkit, in: Proceedings of the
  3rd International Conference on Affective Computing and Intelligent
  Interaction and Workshops (ACII), IEEE, 2009, pp. 1--6.

\bibitem{bib:VerveridisKotropoulos06em}
D.~Ververidis, C.~Kotropoulos, Emotional speech recognition: {Resources},
  features, and methods, Speech Communication 48~(9) (2006) 1162--1181.

\bibitem{bib:WeningerEybenEtAl13acemaud}
F.~Weninger, F.~Eyben, B.~W. Schuller, M.~Mortillaro, K.~R. Scherer, On the
  {Acoustics} of {Emotion} in {Audio}: {What} {Speech}, {Music}, and {Sound}
  have in {Common}, Frontiers in Psychology 4 (May 2013).

\bibitem{haider2018saameat}
F.~Haider, S.~Pollak, E.~Zarogianni, S.~Luz, {SAAMEAT:} active feature
  transformation and selection methods for the recognition of user eating
  conditions, in: Proceedings of the 2018 International Conference on
  Multimodal Interaction (ICMI), ACM, Association for Computing Machinery,
  2018, pp. 564--568.

\bibitem{schuller_interspeech_2015}
B.~Schuller, S.~Steidl, A.~Batliner, S.~Hantke, F.~Hönig, J.~R.
  Orozco-Arroyave, E.~Nöth, Y.~Zhang, F.~Weninger, The {INTERSPEECH} 2015
  computational paralinguistics challenge: nativeness, parkinson's \& eating
  condition, in: Proceedings of the 16th {Annual} {Conference} of the
  {International} {Speech} {Communication} {Association}, INTERSPEECH 2015,
  2015, pp. 478--482.

\bibitem{anagnostopoulos_features_2015}
C.-N. Anagnostopoulos, T.~Iliou, I.~Giannoukos, Features and classifiers for
  emotion recognition from speech: a survey from 2000 to 2011, Artificial
  Intelligence Review 43~(2) (2015) 155--177.

\bibitem{eyben2016geneva}
F.~Eyben, K.~R. Scherer, B.~W. Schuller, J.~Sundberg, E.~Andr{\'e}, C.~Busso,
  L.~Y. Devillers, J.~Epps, P.~Laukka, S.~S. Narayanan, et~al., The geneva
  minimalistic acoustic parameter set (gemaps) for voice research and affective
  computing, IEEE Transactions on Affective Computing 7~(2) (2016) 190--202.

\bibitem{konig_automatic_2015-1}
A.~König, A.~Satt, A.~Sorin, R.~Hoory, O.~Toledo-Ronen, A.~Derreumaux,
  V.~Manera, F.~Verhey, P.~Aalten, P.~H. Robert, Automatic speech analysis for
  the assessment of patients with predementia and {Alzheimer}'s disease,
  Alzheimer's \& Dementia: Diagnosis, Assessment \& Disease Monitoring 1~(1)
  (2015) 112--124.

\bibitem{8910399}
F.~{Haider}, S.~{de la Fuente}, S.~{Luz}, An assessment of paralinguistic
  acoustic features for detection of alzheimer's dementia in spontaneous
  speech, IEEE Journal of Selected Topics in Signal Processing 14~(2) (2020)
  272--281.

\bibitem{goldshtein_automatic_2011-1}
E.~Goldshtein, A.~Tarasiuk, Y.~Zigel, Automatic detection of obstructive sleep
  apnea using speech signals, IEEE Transactions on biomedical engineering
  58~(5) (2011) 1373--1382.

\bibitem{dhall2018emotiw}
A.~Dhall, A.~Kaur, R.~Goecke, T.~Gedeon, Emotiw 2018: Audio-video, student
  engagement and group-level affect prediction, in: Proceedings of the 2018 on
  International Conference on Multimodal Interaction, ACM, 2018, pp. 653--656.

\bibitem{EmotiW2017}
A.~Dhall, R.~Goecke, S.~Ghosh, J.~Joshi, J.~Hoey, T.~Gedeon, From individual to
  group-level emotion recognition: Emotiw 5.0, in: Proceedings of the 19th ACM
  International Conference on Multimodal Interaction (ICMI), ICMI 2017,
  Association for Computing Machinery, 2017, pp. 524--528.

\bibitem{knyazev2017convolutional}
B.~Knyazev, R.~Shvetsov, N.~Efremova, A.~Kuharenko, Convolutional neural
  networks pretrained on large face recognition datasets for emotion
  classification from video, arXiv preprint arXiv:1711.04598 (2017).

\bibitem{Haider:2016:ARV:3011263.3011270}
F.~Haider, L.~S. Cerrato, S.~Luz, N.~Campbell, Attitude recognition of video
  bloggers using audio-visual descriptors, in: Proceedings of the Workshop on
  Multimodal Analyses Enabling Artificial Agents in Human-Machine Interaction,
  MA3HMI 2016, Association for Computing Machinery, 2016, pp. 38--42.

\bibitem{madzlan2015automatic}
N.~A. Madzlan, Y.~Huang, N.~Campbell, Automatic classification and prediction
  of attitudes: Audio-visual analysis of video blogs, in: Proceedings of the
  International Conference on Speech and Computer, Springer, 2015, pp. 96--104.

\bibitem{ICMI2017:HuEtAl}
P.~Hu, D.~Cai, S.~Wang, A.~Yao, Y.~Chen, Learning supervised scoring ensemble
  for emotion recognition in the wild, in: Proceedings of the 19th ACM
  International Conference on Multimodal Interaction (ICMI), ICMI 2017,
  Association for Computing Machinery, 2017, pp. 553--560.

\bibitem{Vielzeuf_Pateux_Jurie_2017}
V.~Vielzeuf, S.~Pateux, F.~Jurie, Temporal multimodal fusion for video emotion
  classification in the wild, in: Proceedings of the 19th ACM International
  Conference on Multimodal Interaction (ICMI), ICMI 2017, Association for
  Computing Machinery, 2017, p. 569–576.

\bibitem{Wang_ICMI2017}
S.~Wang, W.~Wang, J.~Zhao, S.~Chen, Q.~Jin, S.~Zhang, Y.~Qin, Emotion
  recognition with multimodal features and temporal models, in: Proceedings of
  the 19th ACM International Conference on Multimodal Interaction (ICMI), ICMI
  2017, Association for Computing Machinery, 2017, pp. 598--602.

\bibitem{ouyang2017audio}
X.~Ouyang, S.~Kawaai, E.~G.~H. Goh, S.~Shen, W.~Ding, H.~Ming, D.-Y. Huang,
  Audio-visual emotion recognition using deep transfer learning and multiple
  temporal models, in: Proceedings of the 19th ACM International Conference on
  Multimodal Interaction (ICMI), ACM, Association for Computing Machinery,
  2017, pp. 577--582.

\bibitem{hall1999correlation}
M.~A. Hall, Correlation-based feature selection for machine learning, Ph.D.
  thesis, The University of Waikato (1999).

\bibitem{gu_generalized_2012}
Q.~Gu, Z.~Li, J.~Han, Generalized fisher score for feature selection, arXiv
  preprint arXiv:1202.3725 (2012).

\bibitem{wang2006independent}
J.~Wang, C.-I. Chang, Independent component analysis-based dimensionality
  reduction with applications in hyperspectral image analysis, IEEE
  transactions on geoscience and remote sensing 44~(6) (2006) 1586--1600.

\bibitem{jagini2017exploring}
N.~P. Jagini, R.~R. Rao, Exploring emotion specific features for emotion
  recognition system using pca approach, in: Proceedings of the International
  Conference on Intelligent Computing and Control Systems (ICICCS), IEEE, 2017,
  pp. 58--62.

\bibitem{aher2016analysis}
P.~K. Aher, S.~D. Daphal, A.~N. Cheeran, Analysis of feature extraction
  techniques for improved emotion recognition in presence of additive noise,
  in: Proceedings of the International Conference on Computation System and
  Information Technology for Sustainable Solutions (CSITSS), IEEE, 2016, pp.
  350--354.

\bibitem{wang2010speech}
S.~Wang, X.~Ling, F.~Zhang, J.~Tong, Speech emotion recognition based on
  principal component analysis and back propagation neural network, in:
  Proceedings of the International Conference on Measuring Technology and
  Mechatronics Automation (ICMTMA), Vol.~3, IEEE, 2010, pp. 437--440.

\bibitem{Haider2018}
F.~Haider, F.~{A. Salim}, O.~Conlan, S.~Luz,
  \href{http://dx.doi.org/10.21437/Interspeech.2018-1222}{An active feature
  transformation method for attitude recognition of video bloggers}, in: Proc.
  Interspeech 2018, 2018, pp. 431--435.
\newblock \href {https://doi.org/10.21437/Interspeech.2018-1222}
  {\path{doi:10.21437/Interspeech.2018-1222}}.
\newline\urlprefix\url{http://dx.doi.org/10.21437/Interspeech.2018-1222}

\bibitem{roffo2017infinite}
G.~Roffo, S.~Melzi, U.~Castellani, A.~Vinciarelli, Infinite latent feature
  selection: A probabilistic latent graph-based ranking approach, in:
  Proceedings of the IEEE International Conference on Computer Vision (ICCV),
  IEEE, 2017, pp. 1407--1415.

\bibitem{bib:KononenkoSimecRobnik-Sikonja97ovr}
I.~Kononenko, E.~{\v{S}}imec, M.~Robnik-{\v{S}}ikonja, Overcoming the myopia of
  inductive learning algorithms with {ReliefF}, Applied Intelligence 7~(1)
  (1997) 39--55.

\bibitem{kira1992feature}
K.~Kira, L.~A. Rendell, et~al., The feature selection problem: Traditional
  methods and a new algorithm, in: Aaai, Vol.~2, 1992, pp. 129--134.

\bibitem{robnik-sikonja_adaptation_1997}
M.~Robnik-\v{S}ikonja, I.~Kononenko, An adaptation of {Relief} for attribute
  estimation in regression, in: Proceedings of the {Fourteenth} {International}
  {Conference} on Machine Learning ({ICML}), Vol.~5 of ICML 1997, 1997, pp.
  296--304.

\bibitem{kohonen1998self}
T.~Kohonen, The self-organizing map, Neurocomputing 21~(1-3) (1998) 1--6.

\bibitem{burkhardt_database_2005}
F.~Burkhardt, A.~Paeschke, M.~Rolfes, W.~F. Sendlmeier, B.~Weiss, A database of
  german emotional speech, in: Proceedings of the ninth European Conference on
  Speech Communication and Technology, 2005, pp. 1516--1520.

\bibitem{HaqJackson_AVSP09}
S.~Haq, P.~Jackson, Speaker-dependent audio-visual emotion recognition, in:
  Proceedings of the International Conference on Auditory-Visual Speech
  Processing (AVSP), 2009, pp. 53--58.

\bibitem{costantini_emovo_2014}
G.~Costantini, I.~Iaderola, A.~Paoloni, M.~Todisco, Emovo corpus: an italian
  emotional speech database, in: Proceedings of the Ninth International
  Conference on Language Resources and Evaluation (LREC), LREC 2014, European
  Language Resources Association ({ELRA}), 2014, pp. 3501--3504.

\bibitem{eyben2013recent}
F.~Eyben, F.~Weninger, F.~Gro{\ss}, B.~Schuller, Recent developments in
  opensmile, the munich open-source multimedia feature extractor, in:
  Proceedings of the 21st ACM international conference on Multimedia, ACM,
  Association for Computing Machinery, 2013, pp. 835--838.

\end{thebibliography}







\end{document}